\documentclass[a4paper,twocolumn]{scrartcl}\usepackage[]{graphicx}\usepackage[]{color}
\makeatletter
\def\maxwidth{ %
  \ifdim\Gin@nat@width>\linewidth
    \linewidth
  \else
    \Gin@nat@width
  \fi
}
\makeatother

\definecolor{fgcolor}{rgb}{0.345, 0.345, 0.345}

\usepackage{framed}
\makeatletter
 {\par\unskip\endMakeFramed%
 \at@end@of@kframe}
\makeatother

\definecolor{shadecolor}{rgb}{.97, .97, .97}
\definecolor{messagecolor}{rgb}{0, 0, 0}
\definecolor{warningcolor}{rgb}{1, 0, 1}
\definecolor{errorcolor}{rgb}{1, 0, 0}

\usepackage{alltt}      
\usepackage[utf8]{inputenc}
\usepackage[T1]{fontenc}
\usepackage[english]{babel} 
\usepackage{lmodern}
\usepackage[]{algorithm2e}
\usepackage[onehalfspacing]{setspace}
\usepackage{amsmath,amssymb,amsthm,amsfonts,amsbsy,latexsym}
\usepackage{graphicx}
\usepackage{geometry}
\usepackage{natbib}
\usepackage{multirow}
\usepackage{float}
\usepackage{fancyhdr}
\usepackage{makecell}
\usepackage{pgf,tikz}
\usepackage{makeidx}
\usetikzlibrary{shapes,snakes}
\usetikzlibrary{fadings}
\usepackage[pdfpagelabels=true]{hyperref}

\newcommand{\E}{\mbox{I\negthinspace E}}

\newcommand{\R}{\mathbb{R}}

\newcommand{\D}{\mathcal{D}}

\DeclareMathOperator*{\argmax}{argmax}
\DeclareMathOperator*{\rk}{rk}
\DeclareMathOperator*{\sign}{sign}
\DeclareMathOperator*{\Perm}{Perm}

\DeclareMathOperator*{\FPR}{FPR}
\DeclareMathOperator*{\TPR}{TPR}
\DeclareMathOperator*{\IAUC}{IAUC}
\DeclareMathOperator*{\IROC}{IROC}
\DeclareMathOperator*{\ROC}{ROC}
\DeclareMathOperator*{\LocAUC}{LocAUC}

\providecommand{\keywords}[1]
{
  \small	
  \textbf{\textit{Keywords---}} #1
}

\numberwithin{equation}{section}
\IfFileExists{upquote.sty}{\usepackage{upquote}}{}
\begin{document}

\newtheorem{thm}{Theorem}[section]
\newtheorem{Def}{Definition}[section]
\newtheorem{lem}{Lemma}[section]
\newtheorem{rem}{Remark}[section]
\newtheorem{cor}{Corollary}[section]
\newtheorem{ex}{Example}[section]
\newtheorem{ass}{Assumption}[section]
\newtheorem*{bew}{Proof}

\title{A review on instance ranking problems in statistical learning}
\author{Tino Werner\footnote{Institute for Mathematics, Carl von Ossietzky University Oldenburg, P/O Box 2503, 26111 Oldenburg (Oldb), Germany, \texttt{tino.werner1@uni-oldenburg.de}}}
\maketitle

\begin{footnotesize} 

\begin{abstract}

Ranking problems, also known as preference learning problems, define a widely spread class of statistical learning problems with many applications, including fraud detection, document ranking, medicine, credit risk screening, image ranking or media memorability. In this article, we systematically review different types of instance ranking problems, i.e., ranking problems that require the prediction of an order of the response variables, and the corresponding loss functions resp. goodness criteria. We discuss the difficulties when trying to optimize those criteria. As for a detailed and comprehensive overview of existing machine learning techniques to solve such ranking problems, we systemize existing techniques and recapitulate the corresponding optimization problems in a unified notation. We also discuss to which of the ranking problems the respective algorithms are tailored and identify their strengths and limitations. Computational aspects and open research problems are also considered.

\end{abstract} \vspace{0.25cm}


\end{footnotesize}

\begin{small}  

\section{Introduction} 

Search-engines like Google provide a list of web-sites that are suitable for the user's query in the sense that the first web-sites that are displayed are expected to be the most relevant ones. Mathematically spoken, the search-engine has to solve a ranking problem which is done by the \texttt{PageRank} algorithm (\cite{page}) for Google. However, this algorithm is essentially an unsupervised ranking algorithm since it does not invoke any response variable but is based on a graphical model including an adjacency matrix that represents the links connecting the different websites. In this work, we focus on instance ranking problems which belong to the family of supervised ranking problems. \ \\     

In their seminal paper (\cite{clem08}), Cl\'{e}men\c{c}on and co-authors proposed a statistical framework for instance ranking problems which emerge from ordinal regression (\cite{herb}) and proved that the common approach of empirical risk minimization (ERM) is indeed suitable for such ranking problems. Although there already existed instance ranking techniques, most of them indeed follow the ERM principle and can directly be embedded into the framework of \cite{clem08}. \ \\ 

In general, the responses in data sets corresponding to those problems are binary, therefore a natural criterion for such binary ranking problems is the probability that an instance belongs to the class of interest. While ranking can be generally seen in between classification and regression, those binary ranking problems are very closely related to binary classification tasks (see also \cite{balcan08}). For binary ranking problems, there exists vast literature, including theoretical work as well as learning algorithms that use SVMs (\cite{bref}, \cite{herb}, \cite{joachims02}), Boosting (\cite{freund}, \cite{rudinc}), neural networks (\cite{burges}) or trees (\cite{clem08c}, \cite{clem08d}). \ \\

As for the document ranking, the labels may also be discrete, but with $d>2$ classes, for example in the OHSUMED data set (\cite{hersh}). For such general $d-$partite ranking problems, there also has been developed theoretical work (\cite{clem12}), a binary classification approach (\cite{furn09}) as well as tree-based learning algorithms (\cite{clem15}, \cite{clem15b}, see also \cite{robbiano}). \ \\

Recently, Cl\'{e}men\c{c}on investigated a new branch of ranking problems, namely the continuous ranking problems where the name already indicates that the response variable is continuous, with potential applications in natural sciences or quantitative finance (cf. \cite{clem18}). This continuous ranking problem can be located on the other flank of the spectrum of ranking problems that is closest to regression. \ \\

The continuous ranking problem is especially interesting when trying to rank instances whose response is difficult to quantify. A common technique is to introduce latent variables which are used for example to measure or quantify intelligence (\cite{borsboom}), personality (\cite{anand}) or the familiar background (\cite{dickerson}). While in these cases, the latent variables are treated as features, a continuous ranking problem would arise once a response variable which is hard to measure is implicitly fitted by replacing it with some latent score which is much more general than ranking binary responses by means of their probability of belonging to class 1. An example is given in \cite{lan12} where images have to be ranked according to their compatibility to a given query. Another application of continuous ranking problems is given in the risk-based auditing context to detect tax evasion, using the restricted personal resources of tax offices as reasonable as possible. Risk-based auditing can be seen as a general strategy for internal auditing, fraud detection and resource allocation that incorporates different types of risks to be more tailored to the real-world situation, see \cite{pickett} for a broad overview, \cite{moraru} for a short survey of different risks in auditing and \cite{khanna} and \cite{bowlin} for a study on bank-internal risk-based auditing resp. for a study on risk-based auditing for resource planning. \ \\

This paper is organized as follows. Starting in Section \ref{rankingsec} with the definition of several different ranking problems that are distinguished by the shape of the training data, the nature of the response variable and by the goal of the analyst, it becomes evident that suitable loss functions have at least a pair-wise structure in this case. We describe in detail the loss functions corresponding to the different types of ranking problems and related quality criteria which are optimized especially for ranking problems with a discrete response variable. In Section \ref{algsec}, we provide a systematic overview of different machine learning algorithms by grouping them into SVM-, Boosting-, tree- and Neural Network-type approaches. We review these approaches and discuss their strengths, limitations and computational aspects. Section \ref{discuss} is devoted to a careful discussion of the combined ranking problems and a distinguishing between ranking and ordinal regression. We conclude with open research problems for instance ranking.

\section{Supervised ranking problems (instance ranking)} \label{rankingsec} 

\subsection{Different types of ranking problems} 

In order to systematically categorize ranking problems, one has to answer three questions in the following order: What kind of data set do we have (feature-response-pairs, feature-permutation-pairs, only features)? What type of response variable, if it exists, do we have (categorical, continuous)? What is the goal of the analyst? \ \\

At the top level, one can distinguish between label ranking, object ranking and instance ranking problems (cf. \cite{cheng12}). In label ranking problems (see e.g. \cite{har}, \cite{cheng12b}, \cite{furn08}, \cite{huller10b}), the training data consists of features $X_i \in \mathcal{X}$ for some measurable space $\mathcal{X}$ and corresponding permutations $\pi(X_i) \in \Perm(1:d)$ where \begin{center} $ \displaystyle \Perm(1:d):=\{\pi \ | \ \pi \text{ \ is a permutation of \ } \{1,...,d\}\} $ \end{center} denotes the symmetric group on the set $\{1,...,d\}$. A permutation $\pi(X_i)$ is interpreted in the sense that $(\pi(X_i))_1$ represents the most preferred class $c_{(\pi(X_i))_1}$ for instance $X_i$ where $\{c_1,...,c_d\}$ is the set of class labels. Instance ranking considers data consisting of instances $(X_i,Y_i)$ with $Y_i \in \mathcal{Y}$ for some measurable, ordered space $\mathcal{Y}$ where one is interested in finding an ordering of the $X_i$ according to the natural ordering of the $Y_i$ in the respective response space. Object ranking (see e.g. \cite{cohen99}, \cite{huller15}) can be regarded as the unsupervised counterpart of instance ranking, i.e., only features (objects) $X_i$ are given. \ \\

In this review, we only consider instance ranking and we always have data $\D=(X,Y) \in \R^{n \times (p+1)}$ where $Y_i \in \mathcal{Y} \subset \R$ and $X_i \in \mathcal{X} \subset \R^p$ where $X_i$ denotes the $i-$th row of the regressor matrix $X$. \ \\

Solutions of instance ranking problems do not necessarily need to recover the responses $Y_i$ based on the observations $X_i$. In fact, the goal is in general to predict the right ordering of the responses albeit there exist some relaxations of this (hard) ranking problem, e.g., only the top $K<n$ instances have to be ranked exactly while the predicted ranking of the other instances is not a quantity of interest. We go into detail in the next subsection. 

\subsection{Different types of instance ranking problems} 

The goal in this review is to rank the $X_i$ by comparing their predicted response, i.e., $X_i$ will be ranked higher than $X_j$ if $\hat Y_i>\hat Y_j$. We recapitulate the following definitions from \cite{clem08}.  

\begin{Def} \label{rankruledef} \textbf{a)} A \textbf{ranking rule} is a mapping $r: \mathcal{X} \times \mathcal{X} \rightarrow \{-1,1\}$ where $r(x,x')=1$ indicates that $x$ is ranked higher than $x'$ and vice versa.

\textbf{b)} A \textbf{ranking rule induced by a scoring rule $s$} is given by $r(x,x',s)=2I(s(x) \ge s(x'))-1$ with a scoring function $s: \mathcal{X} \rightarrow \R$ where $r(x,x')=1$ if and only if $s(x) \ge s(x')$. \end{Def}

Note that scoring rules are also used in label ranking and object ranking. In label ranking, one prefers class $c_i$ over class $c_j$ for feature $x$ if $s_i(x) \ge s_j(x)$ for scoring functions $s_i,s_j: \mathcal{X} \rightarrow \R$ which in \cite{huller08} are called utility functions. \cite{cohen99} have similar functions in object ranking and refer to them as ordering functions. Furthermore, the ranking rule defined in \cite{clem08} is a hard ranking rule, making a clear decision whether $x$ has to be preferred over $x'$. \cite{cohen99} work with the concept of a probabilistic preference function where a value in $[0,1]$ is assigned to a pair $(x,x')$ where a value close to 1 indicates that $x$ is preferred over $x'$ and vice versa. \ \\

In this work, we will refer to the problem to correctly rank all instances as the \textbf{hard instance ranking problem}\index{Ranking problems!Hard|(} which is a global problem. A weaker problem is the \textbf{localized instance ranking problem}\index{Ranking problems!Localized|(} that intends to find the correct ordering of the best $K$ instances, so misrankings at the bottom of the list are not taken into account. However, misclassifications in the sense that instances that belong to the top $K$ ones are predicted as belonging to the bottom of the list or vice versa have to be additionally penalized in this setting. It is obvious that these two problems are stronger problems than classification problems. \ \\ 

In contrast, sometimes it suffices to tackle the \textbf{weak instance ranking problem}\index{Ranking problems!Weak|(} where one only requires to reliably detect the best $K$ instances but where their pair-wise ordering is not a quantity of interest. 
This problem has been identified in \cite{clem08b} as a classification problem with a mass constraint, since we require to get exactly $K$ class-$1-$objects if class 1 is defined as the ''interesting'' class. We will always denote the index set of the true best $K<n$ instances by $Best_K$ and its empirical counterpart, i.e., the indices of the instances that have been predicted to be the best $K$ ones, by $\widehat{Best_K}$. Worked out theory for the weak and localized instance ranking problem is given in \cite{clem08b}. \ \\

On the other hand, one distinguishes between three other types of instance ranking problems in dependence of the set $\mathcal{Y}$. If $Y$ is binary-valued, w.l.o.g. $\mathcal{Y}=\{-1,1\}$, then a ranking problem that intends to retrieve the correct ordering of the probabilities of the instances to belong to class 1 is called a \textbf{bipartite (binary) instance ranking problem}. If $Y$ can take $d$ different values, a corresponding ranking problem is referred to as a \textbf{$d-$partite instance ranking problem} and for continuously-valued responses, one faces a \textbf{continuous instance ranking problem}. \ \\

So far, we distinguished between different types of ranking problems on different levels. We will discuss in Section \ref{discuss} what combinations define meaningful problems.

\subsection{Loss functions for supervised ranking} 

\subsubsection{Hard ranking} 

Empirical risk minimization requires the definition of a suitable risk function. \cite{clem05} and \cite{clem08} provided the theoretical statistical framework for empirical risk minimization in the ranking setting. The hard ranking risk, i.e., the risk function of the hard instance ranking problem, used in \cite{clem05} and essentially going back to \cite{herb}, is given by \begin{equation} \label{rankrisk} R^{hard}(r):=\E[I((Y-Y')r(X,X')<0)], \end{equation} so in fact, this is nothing but the probability of a misranking of $X$ and $X'$. Thus, empirical risk minimization intends to find an optimal ranking rule by solving the optimization problem \begin{center} $ \displaystyle \min_{r \in \mathcal{R}}(L_n^{hard}(r)) $ \end{center} where  \begin{equation} \label{rankopt} \begin{gathered} L_n^{hard}(r)= \\ \frac{1}{n(n-1)}\mathop{\sum \sum}_{i \ne j} I((Y_i-Y_j)r(X_i,X_j)<0) \end{gathered} \end{equation} where $\mathcal{R}$ is some class of ranking rules $r: \mathcal{X} \times \mathcal{X} \rightarrow \{-1,1\}$. For the sake of notation, the additional arguments in the loss function are suppressed. Note that $L_n^{hard}$, i.e., the hard empirical risk, is also the hard ranking loss and not a sum of individual instance-wise losses as in regression or classification settings which reflects the global nature of hard ranking problems.  \ \\

In the instance ranking setting, ranking rules induced by scoring rules are self-evident due to the natural ordering existing on $\mathcal{Y}$. Considering some parameter space $\Theta \subset \R^p$, it suffices to empirically find the best parametric scoring function (and with it, the empirically optimal induced ranking rule) from the family \begin{center} $ \displaystyle \mathcal{S}:=\{s_{\theta}: \mathcal{X} \rightarrow \R \ | \ \theta \in \Theta\}$ \end{center} of such scoring functions by solving the parametric optimization problem \begin{center} $ \displaystyle \min_{\theta \in \Theta}(L_n^{hard}(\theta)) $ \end{center} with \begin{equation} \label{rankparopt} \begin{gathered} L_n^{hard}(\theta)=\frac{1}{n(n-1)} \\ \mathop{\sum \sum}_{i \ne j} I((Y_i-Y_j)(s_{\theta}(X_i)-s_{\theta}(X_j))<0). \end{gathered} \end{equation} 

We insist to once more take a look on the U-statistics that arise for the hard and the localized ranking problem. \cite{clem08} already mentioned that these pair-wise loss functions can be generalized to loss functions with $m$ input arguments. This leads to U-statistics of order $m$. But if the whole permutations that represent the ordering of the response values should be compared at once (i.e., $m=n$), then this again boils down to a U-statistic of order 2. Let $\pi$, $\hat \pi \in \Perm(1:n)$ be the true resp. the estimated permutation, then the empirical hard ranking loss can be equivalently written as\begin{equation}   \label{rankriskperm} \begin{gathered} L_n^{hard}(\pi, \hat \pi)=\\ \frac{2}{n(n-1)}\mathop{\sum \sum}_{i < j} I((\pi_i-\pi_j)(\hat \pi_i-\hat \pi_j)<0). \end{gathered} \end{equation} In fact, this loss function can be identified with the ranking loss used in \cite{huller17} in the context of object ranking where the training data consists of sets of features including the corresponding true permutations representing the ordering on the respective subset. 

\subsubsection{Weak ranking} 

For the weak instance ranking problem, \cite{clem08b} introduce the upper $(1-u)-$quantile $Q(s,1-u)$ for the random variable $s(X)$ for binary responses. Since a weak ranking problem can also be formulated for continuous-valued responses, we consider the transformed responses \begin{center} $ \displaystyle \tilde Y_i^{(K)}:=2I(\rk(Y_i) \le K)-1 $ \end{center} where the ranks come from a descending ordering, i.e., \begin{center} $ \displaystyle \rk(Y_i)=\sum_j I(Y_i \le Y_j) $ .\end{center} Then the misclassification risk corresponding to the weak instance ranking problem in the sense of \cite{clem08b} is given by  \begin{center} $ \displaystyle R^{weak,u}(s):=P(\tilde Y(s(X)-Q(s,1-u))<0) $ \end{center}  with the empirical counterpart  \begin{equation*} \begin{gathered}  L_n^{weak, K}(s)= \\ \frac{1}{n}\sum_{i=1}^n I(\tilde Y_i^{(K)}(s(X_i)-\hat Q(s,1-u^{(K)}))<0)  \end{gathered}  \end{equation*}  for the empirical quantile $\hat Q(s,1-u^{(K)})$. To approximate the $(1-u)-$quantile, one needs to set $u^{(K)}=K/n$, i.e., for a given level $(1-u)$, one looks at the top $K$ instances that represent this upper quantile. 

\begin{rem} Due to the mass constraint, each false positive generates exactly one false negative, so the loss can be equivalently written as  \begin{equation*} \begin{gathered} L_n^{weak, K}(s)= \\ \frac{2}{n}\sum_{i \in Best_K} I(\tilde Y_i^{(K)}(s(X_i)-\hat Q(s,1-u^{(K)}))<0). \end{gathered} \end{equation*} \end{rem} 

Note that the weak ranking loss is not standardized, i.e., it is not necessarily able to take the value 1. More precisely, its maximal value is always $\frac{2K}{n}$, so we can only hit the value one if $K=\frac{n}{2}$ for even $n$ and if all instances that belong to the ''top half'' and predicted to be in the ''bottom half'' and vice versa. For better comparison of the losses, \cite{TWphd} propose the \textbf{standardized weak ranking loss} \begin{equation} \label{weakranknorm} \begin{gathered} L_n^{weak, K, norm}(s)=\frac{1}{K} \\ \sum_{i \in Best_K} I(\tilde Y_i^{(K)}(s(X_i)-\hat Q(s,1-u^{(K)}))<0). \end{gathered} \end{equation} 

\begin{rem} Having get rid of the ratio $K/n$, the standardized weak ranking loss function has a very intuitive interpretation. For a fixed $K$, a standardized weak ranking loss of $c/K$ means that $c$ of the instances of $Best_K$ did not have been recovered by the model. \end{rem} 

\subsubsection{Localized ranking} 
                
A suitable loss function for the localized instance ranking problem was proposed in \cite{clem08b}, too. In our notation, it is given by \begin{equation} \label{locrankrisk} \begin{gathered} L_n^{loc, K}(s):=\frac{n_-}{n} L_n^{weak, K}(s)+\frac{1}{n(n-1)}\mathop{\sum \sum}_{i \ne j} \\ I(\{(s(X_i)-s(X_j))(Y_i-Y_j)<0\} \\ \cap\{\min(s(X_i),s(X_j)) \ge \hat Q(s,1-u^{(K)})\})  \end{gathered} \end{equation} In the first summand, $n_-$ indicates the number of negatives, so the quotient is just an estimate for $P(Y=-1)$. Note that \cite{clem08b} introduced this loss for binary-valued responses. We propose to set $n_-:=(n-K)$ for continuously-valued responses since localizing artificially labels the top $K$ instances as class 1 objects, hence we get $(n-K)$ negatives. Again, the second summand may be rewritten as \begin{center} $ \displaystyle \frac{2}{n(n-1)}\mathop{\sum \sum}_{i<j, i, j \in \widehat{Best_K}} I((s(X_i)-s(X_j))(Y_i-Y_j)<0). $ \end{center} As the weak ranking loss, this loss is not $[0,1]-$standardized. Taking a closer look on it, the maximal achievable loss given a fixed $K$ is \begin{center} $ \displaystyle \max(L_n^{loc,K}(s))=\frac{n-K}{n} \cdot \frac{2K}{n}+\frac{K(K-1)}{n(n-1)}=:m_K, $ \end{center} so a standardized version is simply \begin{center} $ \displaystyle L_n^{loc,K,norm}(s):=\frac{1}{m_K}L_n^{loc,K}(s). $ \end{center} 

\begin{rem} Note that even in the case $K=\frac{n}{2}$ for even $n$, the localized ranking loss cannot take the value one. This is true since \begin{center} $ \displaystyle L_n^{loc,n/2}(s) \le \frac{\frac{n}{2}}{n}+\frac{\frac{n}{2}\left(\frac{n}{2}-1\right)}{n(n-1)} \cdot 1<\frac{1}{2}+\frac{\frac{1}{2}n(n-1)}{n(n-1)}=1. $ \end{center} \end{rem} 

A simple example for clarification is given below in Example \ref{ranklossex} which we borrow from \cite{TWphd}. 

\begin{ex} \label{ranklossex} Assume that we have a data set with the true response values \begin{center} $ \displaystyle Y:=(-3,10.3,-8,12,14,-0.5,29,-1.1,-5.7,119) $ \end{center} and the fitted values $\hat Y$ given by \begin{center} $ \displaystyle (0.02,0.6,0.1,0.47,0.82,0.04,0.77,0.09,0.01,0.79). $ \end{center} Then we order the vectors according to $Y$, so that $Y_1 \ge Y_2 \ge ...$ and get the permutations \begin{center} $ \displaystyle \pi=(1,2,...,10), \ \ \ \hat \pi=(2,3,1,5,4,8,7,9,10,6). $ \end{center} For example, $Y_{10}=119$ is the largest value of $Y$, having rank 1. So we reorder $\hat Y$ such that $\hat Y_{10}=0.79$ is the first entry. But since this is only the second-largest entry of $\hat Y$, we have a rank of 2, leading to the first component $\hat \pi_1=2$ and so forth. \ \\

Setting $K=4$, we obviously get \begin{center} $ \displaystyle L_n^{weak,4}(\pi, \hat \pi)=\frac{2}{10}=0.2. $ \end{center} The standardized weak ranking loss\index{Ranking losses!Standardized weak|)} is then \begin{center} $ \displaystyle L_n^{weak,4,norm}(\pi, \hat \pi)=\frac{10}{8} \cdot \frac{2}{10}=0.25 $ \end{center} which is most intuitive since one of the indices of the four true best instances is not contained in the predicted set $\widehat{Best_4}$. The second part of the localized loss is then \begin{center} $ \displaystyle \frac{2}{90}[0+1+0+1+0+0]=\frac{2}{45}. $ \end{center} This makes it obviously why the misclassification loss has to be included since this loss would be same if the instances of rank 4 and 5 were not switched. The complete localized ranking loss is \begin{center} $ \displaystyle L_n^{loc,4}(\pi, \hat \pi)=\frac{2}{45}+\frac{6}{10} \cdot 0.2=\frac{37}{225}. $ \end{center} The standardized localized ranking loss is then \begin{center} $ \displaystyle L_n^{loc,4,norm}(\pi, \hat \pi)=\frac{75}{46}\cdot \frac{37}{225} \approx 0.268. $ \end{center} Finally, the hard ranking loss is \begin{center} $ \displaystyle L_n^{hard}(\pi, \hat \pi)=\frac{2}{90} \cdot 8=\frac{16}{90}. $\end{center} 

Setting $K=5$, the weak ranking loss\index{Ranking losses!Weak|)} is zero and the localized ranking loss\index{Ranking losses!Localized|)} is \begin{center} $ \displaystyle L_n^{loc,5}(\pi, \hat \pi)=\frac{2}{90}[0+1+0+0+1+0+0+0+0+1]+\frac{5}{10} \cdot 0=\frac{1}{15}. $ \end{center} The standardized localized ranking loss\index{Ranking losses!Standardized localized|)} is \begin{center} $ \displaystyle L_n^{loc,5,norm}(\pi, \hat \pi)=\frac{18}{13} \cdot \frac{1}{15} \approx 0.092. $ \end{center} The hard ranking loss\index{Ranking losses!Hard|)} is a global loss and does not change when changing $K$. \ \\

This nice and simple example has shown how important the selection of $K$ can be. \end{ex}

\subsection{Fast computation of the hard ranking loss} 

A na\"{i}ve evaluation of the hard ranking loss requires $\mathcal{O}(n^2)$ comparisons. This will surely become infeasible for data sets with many observations, therefore \cite{TWphd} provided a solution which only requires $\mathcal{O}(n\ln(n))$ evaluations. \ \\

They take a look at the concordance measure Kendall's $\tau$, i.e., \begin{center} $ \displaystyle \tau(Y,\hat Y):=\frac{1}{n(n-1)}\mathop{\sum \sum}_{i \ne j} \sign((Y_i-Y_j)(\hat Y_i-\hat Y_j)) $. \end{center} Unlike the ranking loss which is high if there are many misrankings and which is $[0,1]-$valued, Kendall's $\tau$ is high if many pairs are concordant, i.e., if the pair-wise ranking is correct in most cases and takes values in $[-1,1]$. \ \\

This leads to a bijection between these two quantities if we do not face ties as given in the following lemma from \cite{TWphd}. 

\begin{lem}[\textbf{Hard ranking loss and Kendall's Tau}] \label{fasthardrank} Assume the vectors $y$ and $y'$ have the same length $n$ and do not contain ties. Then it holds that \begin{center} $ \displaystyle L_n^{hard}(y,y')=\frac{1-\tau(y,y')}{2}. $ \end{center} \end{lem} 

This indeed turns out to be useful in practice since there exists a fast computation method for Kendall's $\tau$ essentially going back to Knight (\cite{knight66}) which relies on the idea of fast ordering algorithms and which is implemented for example as \texttt{cor.fk} in the $\mathsf{R}-$package \texttt{pcaPP} (\cite{pcapp}). The algorithm reduces the number of calculations necessary for the computation of the hard ranking loss from $\mathcal{O}(n^2)$ in the na\"{i}ve implementation to $\mathcal{O}(n\ln(n))$.

\subsection{Quality criteria for ranking} 

So far, we presented loss functions for instance ranking problems that lead to algorithms in the spirit of the ERM (later also the SRM) paradigm. On the other hand, there also exist quality measures that are popular in classification settings but which already have been transferred to the ranking setting. Before we go into detail, we recapitulate the definition of a common and well-known quality criterion for classification. 

\begin{Def} \label{rocauc} Let $Y_1,...,Y_n$ take values in $\{-1,1\}$ where the total number of positives is $n_+$ and the total number of negatives is $n_-$. Let $\hat Y_i \in \{-1,1\}$, $i=1,...,n$, be predicted values. 

\textbf{a)} The \textbf{true positive rate (TPR)} and the \textbf{false positive rate (FPR)} are given by  \begin{center} $ \displaystyle \TPR=\frac{1}{n_+}\sum_i I(\hat Y_i=1)I(Y_i=1),  $ \end{center} \begin{center} $ \displaystyle \FPR=\frac{1}{n_-}\sum_i I(\hat Y_i=1)I(Y_i=-1). $ \end{center} 

\textbf{b)} The \textbf{Receiver Operation Characteristic curve (ROC curve)} is the plot of the true positive rate against the false positive rate. 

\textbf{c)} The AUC is defined as the \textbf{area under the ROC curve}. \end{Def}  

For theoretical aspects of the empirical AUC and its optimization, we refer to \cite{agar}, \cite{cortes} and \cite{calders}. We continue presenting the reparametrization of the ROC curve as it has been introduced in \cite{clem08} and used in subsequent papers of Cl\'{e}men\c{c}on and coauthors.  

\begin{Def} For a scoring function $s$, the true positive rate and the false positive rate are given by \begin{equation*} \begin{gathered} \TPR_s(x)=P(s(X) \ge x \ | \ Y=1) \\ \FPR_s(x)=P(s(X) \ge x \ | \ Y=-1)  \end{gathered}. \end{equation*} Setting \begin{center} $ \displaystyle q_{s,\alpha}:=\inf\{x \in ]0,1[ \ | \ \FPR_s(x) \le \alpha\}, $ \end{center} the ROC curve is the plot of $\TPR_s(q_{s, \alpha})$ against the level $\alpha$. \end{Def} 

The ROC curve is a standard tool to validate binary classification rules. If the classification depends on a threshold, different points of the ROC curve are generated by changing the threshold and computing the TPR and the FPR. Since the goal is to achieve a TPR as high as possible for the price of an FPR as low as possible, one usually chooses the threshold corresponding to the upper-leftmost point of the empirical ROC curve. A combined quality measure that incorporates all points of the ROC curve is the AUC where a classification rule is better the higher the empirical AUC is. Random guessing clearly has a theoretical AUC of 0.5. \ \\

For the bipartite localized ranking problem, \cite{clem08b} provide the following localized version of the AUC. It is important to note a strong equivalence between the AUC and the ranking error $P((Y-Y')(s(X)-s(X'))<0)$ in the sense that minimizing this error is equivalent to maximizing the AUC corresponding to the scoring function $s$ (see \cite{clem08b}). 

\begin{Def}\label{locaucdef} The \textbf{localized AUC} is defined as \begin{equation*} \begin{gathered} \LocAUC(s,\alpha):=P(\{s(X)>s(X')\} \cap \\ \{s(X) \ge Q(s,1-\alpha)\} \ | \ Y=1, Y'=-1) \end{gathered}. \end{equation*} \end{Def}  

As for $d-$partite ranking problems, i.e., $Y$ can take $d$ different values, w.l.o.g. $\mathcal{Y}=\{1,...,d\}$ with ordinal classes, \cite{clem12} proposed the VUS (volume under the ROC surface) as quality criterion. 

\begin{Def} \label{vusdef} Let w.l.o.g. $Y$ take values in $\{1,...,d\}$ and let again $X$ take values in $\mathcal{X} \subset \R^p$. For a scoring function $s: \mathcal{X} \rightarrow \R$, define \begin{center} $ \displaystyle F_{s,k}(t):=P(s(X) \le t|Y=k) $ \end{center} for $k=1,...,d$. \\
\textbf{a)} The ROC surface is the ''continuous extension'' (\cite{clem12}) of the plot \begin{center} $ \displaystyle (t_1,...,t_{d-1}) \mapsto (F_{s,1}(t_1), F_{s,2}(t_2)-F_{s,2}(t_1),...,1-F_{s,d}(t_{d-1})) $ \end{center} for $t_1<t_2<...<t_{d-1}$. \\
\textbf{b)} The VUS is the volume under the ROC surface. \end{Def} 

In this definition, the term ''continuous extension'' means to connect the points by hyperplane parts as described in \cite{clem12}. The ROC surface can be interpreted as joint plot of the class-wise true positive rates since if the value of the scoring function is between $t_k$ and $t_{k+1}$ (artificially define $t_0:=-\infty$ and $t_d:=\infty$), the instance is assigned to class $(k+1)$. \\

The VUS is not the only possible way how to assess the quality of multipartite ranking models. \cite{furn09} considered the C-index \begin{center} $ \displaystyle C(s,X)=\frac{1}{\sum_{i<j} n_in_j} \mathop{\sum \sum \sum}_{i<j, (x,x') \in C_i \times C_j} I(s(x')>s(x)) $ \end{center} where $C_i$ denotes the set of all class-$i-$instances with $n_i=|C_i|$, measuring the probability that a randomly selected class-$j-$instance is (correctly) ranked above a randomly chosen class-$i-$instance, and the extension \begin{center} $ \displaystyle U(s,X)=\frac{2}{d(d-1)}\mathop{\sum \sum}_{i<j} AUC(s,C_i \cup C_j) $, \end{center} of the AUC which in \cite{furn09} is identified with a weighted version of the C-index. \cite{waegeman} proposed the metric \begin{equation*} \begin{gathered} W(f(X))=\\ \frac{1}{\prod_i n_i}  \sum_{X_1 \in C_1,...,X_d \in C_d} I(s(X_1)<...<s(X_d)) \end{gathered} \end{equation*} which however, as discussed in \cite{furn09}, neglects how severely the ordering is violated, i.e., if there is only one misranking between two of the $d$ instances or if the ordering is even reverted. \cite{furn09} therefore concentrate on $C(s,X)$ and $U(s,X)$. \ \\

Other well-known quality criteria for ranking problems are for example the MAP (mean average precision) and the NCDG (normalized discounted cumulative gain) which are mainly used in object ranking (see \cite{cheng10b}).

\section{Current techniques to solve ranking problems} \label{algsec} 

This section is divided into four parts. Each subsection is devoted to a particular underlying machine learning algorithm for the discussed ranking approach, i.e., Support Vector Machines (SVM), Boosting, trees and Neural Networks resp. Deep Learning.

\subsection{SVM-type approaches} 

\cite{joachims02} provided the \texttt{RankingSVM} algorithm for document retrieval which is essentially based on the seminal approach for ordinal regression introduced in \cite{herb}. In the situation of \cite{herb}, a set of pairs $(X_i,Y_i)$ is given. Their goal is to solve a hard bipartite ranking problem, but they do not optimize the hard ranking loss directly but formulate the constraint inequalities in the sense that $s(X_i)>s(X_j)$ for $X_i$ being more relevant than $X_j$, given each of the queries. As \cite{joachims02} argue, trying to find a scoring function such that every inequality is satisfied would be NP-hard, so \cite{herb}, \cite{joachims02} introduce slack variables and formulate the problem as a standard SVM problem but with all the relaxed inequalities as constraints, so that one gets a standard SVM-type solution $s(x)=\sum_i \alpha_i K(x,X_i)$ for a kernel $K$ (\cite{herb}). Due to the equivalence of SVM problems and structural risk minimization problems with a Hinge loss, \cite{clem13} translated the criterion in \cite{joachims02} into the regularized pair-wise empirical loss \begin{center} $ \displaystyle \frac{2}{n(n-1)} \mathop{\sum \sum}_{i<j} [1-(Y_i-Y_j)(s(X_i)-s(X_j))]_++\lambda ||s||_{\mathcal{H}_K}^2 $ \end{center} where $\mathcal{H}_K$ is some Reproducing Kernel Hilbert Space (RKHS) defined by a kernel $K$ (see for example \cite{schol01}). They call their algorithm \texttt{RankingSVM}.  Note that \cite{joachims02} essentially have a data set consisting of documents and queries. The goal is that for a given query, a scoring function $s$ has to be computed such that the ordering of the documents according to the scoring function is as concordant as possible with the true ordering according to the relevance of the documents w.r.t. the query. They point out that this setting is more flexible than the one in \cite{herb} since it allows for different rankings for different queries. \ \\

An implementation of \texttt{RankingSVM} is given in the SVM${}^{light}$ software package (\cite{joachims99a}) in \texttt{C} language \footnote{http://svmlight.joachims.org/} as well as an improved implementation in the software package SVM${}^{rank}$ relying on the cutting-plane algorithm from \cite{joachims}. As for the computation of the solutions, note that \cite{chapelle09} argued that the SVM${}^{light}$ implementation for \texttt{RankingSVM} requires the computation of all pairwise differences $X_i-X_j$ which leads to a complexity of $\mathcal{O}(n^2)$. They propose a truncated Newton step which is computed via conjugate gradients in order to remedy this issue and result with the MATLAB implementation \texttt{PRSVM} \footnote{http://olivier.chapelle.cc/primal/}, essentially reducing the respective complexity to $\mathcal{O}(np)$ for $n>p$. \cite{chen17} accelerate the computation of the kernel matrix for the case $n>p$ by invoking the kernel approximation $K(x,x')=\langle \Phi(x), \Phi(x') \rangle$ which generates an approximate kernel Hilbert space and provides an SVM solution of the form \begin{center} $ \displaystyle s(x)=w^T\Phi(x). $ \end{center} They propose two methods to get a suitable kernel approximation. The first is a Nyström approximation where $m\ll n$ rows of $X$, say, $\hat X_1$,...,$\hat X_m$, are sampled uniformly, followed by a singular value decomposition of the matrix $(K(\hat X_i,\hat X_j))_{i,j=1,...,m}$. Truncating the SVD by taking just the first $k$ columns of the orthonormal matrix and the upper left $k \times k-$submatrix of the diagonal matrix, one gets a rank-$k-$approximation, reducing the complexity to $\mathcal{O}(npk+k^3)$. Another strategy is to Fourier transform the kernel, i.e., \begin{center} $ \displaystyle K(x,x')=\int q(\omega)exp(i\omega^T(x-x'))d\omega $, \end{center} and to draw $m$ samples according to $q$, providing a kernel approximation using Bochner's theorem. Despite the approximation error is higher than for the Nyström approximation (for equal $m$), the complexity is just $\mathcal{O}(nmp)$. \cite{chen17} provide publicly available MATLAB code \footnote{https://github.com/KaenChan/rank-kernel-appr}. \ \\

\cite{rak04} and \cite{ataman} use the fact that the binary hard ranking problem can be solved by maximizing the AUC of the scoring function. Since the responses are binary-valued, \cite{rak04} explicitely distinguishes between positive and negative instances by writing $X_i^+$ resp. $X_i^-$ for the features. The empirical AUC can be estimated by \begin{center} $ \displaystyle \widehat{AUC}=\frac{1}{n_-n_+}\sum_{i=1}^{n_-}\sum_{j=1}^{n_+} I(s(X_i^+)>s(X_j^-))=\frac{1}{n_-n_+}\sum_{i=1}^{n_-}\sum_{j=1}^{n_+} I(\xi_{ij}>0) $ \end{center} for $\xi_{ij}:=s(X_i^+)-s(X_j^-)$. Using this definition of $\xi_{ij}$ as equality constraint, \cite{rak04} formulate the problem as SVM-type problem by considering linear scoring functions $s(x):=w^Tx+b$. They show that the solution essentially has the form \begin{center} $ \displaystyle s(x)=\sum_i \sum_j \alpha_{i,j} (K(X_i^+,x)-K(X_j^-,x))+b. $ \end{center} in the general case when using kernels. In \cite{rak04}, the algorithm is applied to different data sets, including a cancer and a credit data set. They conclude that their algorithm also provides good accuracy performances. \cite{ataman} used a MATLAB and a WEKA implementation and the algorithm from \cite{rak04} can be found in a MATLAB toolbox \footnote{http://asi.insa-rouen.fr/enseignants/~arakoto/toolbox/} (\cite{rakmat}). \ \\

\cite{bref} provide a very similar approach, but they both provide a so-called ''1-Norm'' and ''2-Norm'' problem, namely \begin{center} $ \displaystyle \frac{1}{2}||w||^2+\frac{C}{2}\sum_i \sum_j \xi_{ij}^r $ \end{center} for the target function of the SVM, where $r \in \{1,2\}$, and the corresponding solutions. A recommendation for the choice of $r$ is however not given. Due to the evaluation of the kernel matrix and the quadratically growing number of constraints, their algorithm is of complexity $\mathcal{O}(n^4)$. They provide some suggestions how to reduce the complexity. \ \\

\cite{cao} argue that a major weakness of \texttt{RankingSVM} is that misrankings on the top of the list get the same loss as misrankings at the bottom. Therefore, they propose a weighted variant of the Hinge loss in the sense that the weights are higher the higher the importance of the documents and the queries is. They apply their algorithm to the OHSUMED data set.  \ \\

\cite{jung} provide \texttt{Ensemble RankingSVM} by combining different \texttt{RankingSVM} models. \ \\

Since SVM-type solutions are not sparse, there are several approaches to construct SVM-type ranking functions with feature selection. \ \\

\cite{tian} consider essentially the same problem as \cite{rak04}, but with the crucial difference that the target function is \begin{center} $ \displaystyle ||w||_q^q+C\sum_i \sum_j \xi_{ij}$ \end{center} for $0<q<1$, so $||w||_2^2$ has been replaced by a concave loss. They solve the problem with a multi-stage convex relaxation technique. They conclude that by the $l_q-$norm, the algorithm indeed performs feature selection which results from the equivalence to write an SVM problem as a regularized problem with the Hinge loss. Since the number of constraints grows quadratically with the number of observations, they propose to cluster the observations first and to just perform the computations on the representants. \ \\

Another approach is given in \cite{lai13} where they replace the quadratic penalty (i.e., $||w||_2^2$ in the equivalent formulization) with an $l_1-$regularization term and use the squared Hinge loss. They solve the problem by invoking Fenchel duality (hence the name \texttt{FenchelRank}) and prove convergence of the solution. After experiments on real data sets for document retrieval, they conclude sparsity of the solutions as well as superiority of \texttt{FenchelRank} to non-sparse algorithms. They implement their method in MATLAB. An iterative gradient procedure for this problem has been developed in \cite{lai13b} and shows comparable performance. \ \\

As an extension of \texttt{FenchelRank}, \cite{laporte} tackle the analogous problem with nonconvex regularization to get even sparser models. They solve the problem with a majorization minimization method where the nonconvex regularization term is represented by the difference of two convex functions. In addition, for convex regularization, they present an approach that relies on differentiability and Lipschitz continuity of the penalty term so that the \texttt{ISTA}-algorithm can be applied. They provide publicly available MATLAB code \footnote{http://remi.flamary.com/soft/soft-ranksvm-nc.html}. \ \\ \ \\

Another approach that does not provide an SVM-type solution at the first glance is given in \cite{pahi}. They intend to predict the differences of the responses by the differences of the scores assigned to the respective features, i.e., to essentially solve  \begin{equation*} \begin{gathered} \frac{1}{n(n-1)} \mathop{\sum \sum}_{i<j} \frac{1}{2}|\sign(s(X_i)-s(X_j))-\sign(Y_i-Y_j)| \\ +\lambda ||s||_{\mathcal{H}_K}^2  \end{gathered} \end{equation*} for some kernel $K$ with corresponding RKHS $\mathcal{H}_K$.  Since this problem is clearly not tractable, as \cite{pahi} point out, they instead minimize the regularized least-squares-type criterion \begin{equation*} \begin{gathered} \frac{1}{n(n-1)} \mathop{\sum \sum}_{i<j} ((s(X_i)-s(X_j))-(Y_i-Y_j))^2 \\ +\lambda ||s||_{\mathcal{H}_K}^2 \end{gathered} \end{equation*} Using the representer theorem (see e.g. \cite{schol01}), the solution has the form  \begin{center} $ \displaystyle f(X)=\sum_{i=1}^n \alpha_iK(X,X_i) $ \end{center} for some $\alpha_i \in \R$. The algorithm is called \textbf{RankRLS} (''regularized least squares''). The complexity of the algorithm is of order $\mathcal{O}(p^3+np^2)$ resulting from matrix inversion and matrix multiplication. Note that \cite{pahi10} provided a greedy method to compute the respective inverse by successively selecting up to $k<p$ features which results in an overall complexity of their \texttt{greedy RankRLS} algorithm of $\mathcal{O}(knp)$. \cite{pahi10} provided a link leading to implementations of both \texttt{RankRLS} and \texttt{gree\-dy RankRLS}, but it does not seem to be available anymore. \ \\ 

Summarizing, there exist a rich variety of SVM-type ranking algorithms in order to minimize the hard ranking loss, including approaches that provide sparse solutions. The approach of \cite{cao} minimizes a weighted hard ranking loss and can be seen as the closest SVM-type approach for localized ranking problems. In general, these SVM-type ranking algorithms are tailored to bipartite ranking problems. Note that SVM solutions are in general hard to interpret. In contrast to the AUC-maximizing approaches, the other algorithms make use of a surrogate loss function for the hard ranking loss which is either a pair-wise Hinge or pair-wise squared loss.

\subsection{Boosting-type approaches} 

In the case of bipartite ranking, the sometimes called ''plug-in approach'' that estimates the conditional probability $P(Y=1|X=x)$ can be realized for example by \texttt{LogitBoost} (see e.g. \cite{bu}), i.e., minimizing the loss \begin{center} $ \displaystyle \frac{1}{n} \sum_i \log_2(1+\exp(-2Y_is(X_i))).$ \end{center} The resulting function $s$ is then used as a ($[0,1]-$valued) scoring function for the ranking. However, the plug-in approach has disadvantages when facing high-dimensional data and it furthermore just optimizes the ROC curve in an $L_1-$sense as pointed out in \cite{clem08c}, \cite{clem08d}. Taking a closer look on this loss function, it is indeed a convex surrogate of the misclassification loss and does not respect a pair-wise structure. Concerning informativity, one just applies an algorithm that solves a classification problem which is less informative than a ranking problem (see also \cite{furn09}) which is another aspect why this approach cannot be optimal. As mentioned in \cite{clem13}, a kernel logistic regression may also be thinkable in the same plug-in sense (which has the same weaknesses). \ \\

\cite{freund} developed a Boosting-type algorithm (\texttt{RankBoost}) which combines weak rankers in an \texttt{AdaBoost}-style (for the latter, see \cite{freund97}) benefitting from the binarity of the response variable. First, they propose a distribution $D$ on the space $\mathcal{X} \times \mathcal{X}$ which, for data $\D$, is represented as a matrix that essentially contains weights. These weights can be thought of representing the importance to rank the corresponding pair correctly. As for the weak rankers which are nothing but a scoring function $\tilde s$, they consider either the identity function or a function that maps the features essentially into the set $\{0,1\}$ according to some threshold. More precisely, the weak ranker is chosen such that the quality measure \begin{center} $ \displaystyle  \mathop{\sum \sum}_{i \ne j: r(X_i,X_j)=1} D(X_i,X_j)(\tilde s(X_i)-\tilde s(X_j)) $ \end{center} is maximized where $r$ again denotes a ranking rule as introduced in Definition \ref{rankruledef}, meaning that the sum runs over all pairs $(X_i,X_j)$ where $X_i$ is ranked higher than $X_j$. As the \texttt{AdaBoost} algorithm minimizes the exponential surrogate of the 0/1-classification loss, \cite{clem13} pointed out that \texttt{RankBoost} minimizes the pair-wise surrogate loss function \begin{center} $ \displaystyle \frac{1}{n(n-1)}\mathop{\sum \sum}_{i<j} \exp(-(Y_i-Y_j)(s(X_i)-s(X_j))) $. \end{center} Note that there is a small mistake in Section 3.2.1 of \cite{clem13} since the minus sign in the exponential function is missing. \ \\

It is shown in \cite{rudin09} that in the case of binary outcome variables, \texttt{RankBoost} and the classifier \texttt{AdaBoost} are equivalent under very weak assumptions. Therefore, \texttt{RankBoost} can also be seen as an AUC maximizer in the bipartite ranking problem. \cite{freund} apply \texttt{RankBoost} for document retrieval. The algorithm is available at the RankLib library (\cite{ranklib}). \ \\

An extension of \texttt{RankBoost} has been provided in \cite{rudinc}. They intend to optimize essentially \begin{center} $ \displaystyle \frac{2}{n(n-1)} \sum_i \left( \mathop{\sum}_{j>i} \exp((Y_i-Y_j)(s(X_i)-s(X_j))) \right)^p $ \end{center} for some $p \ge 1$ (\cite{rudinc} originally distinguish positive and negative instances, but \cite{clem13} used the notation as in the display above). The argument behind this power loss given in \cite{rudinc} is that the higher $p$ is chosen, the higher the difference between the loss of misrankings at the top of the list and misrankings at the bottom of the list becomes. The algorithm parallels the \texttt{RankBoost} algorithm in combining weak rankers, but since the weights are not always analytically computable, they may use a linesearch. They call their algorithm \texttt{p-Norm-Push}. The case $p=\infty$ has been studied in \cite{rak12}. \ \\

So, while \texttt{RankBoost} is tailored to hard bipartite instance ranking problems (and may be used for $d-$partite instance ranking problems in the sense of \cite{clem12}), the \texttt{p-Norm-Push} is closest to handle localized bipartite instance ranking problems. However, the results of the simulation study in \cite{clem13} reveal that the localized AUC criterion for the corresponding predictions is not better than for \texttt{RankBoost}. To the best of our knowledge, the \texttt{p-Norm-Push} has never been applied to $d-$partite ranking problems. \ \\ 

Another generalization of \texttt{RankBoost} has been proposed in \cite{zheng}, again for hard bipartite instance ranking problems. They suggest to use a sufficiently regular surrogate of the ranking loss like a squared or a squared Hinge loss and to apply Gradient Boosting (\cite{friedman01}, \cite{bu07}) to this surrogate loss.  As weak learner, they consider a so-called ''regression weak learner'' to fit the gradients in each iteration. They apply their algorithm to document retrieval data. \ \\ 

In contrast to the already reviewed Boosting-type approa\-ches which are designed for bipartite instance ranking problems, \cite{TWphd} argue that in the context of risk-based auditing (see e.g. \cite{alm}, \cite{gupta}, \cite{hsu15}), it is more reasonable to solve a continuous ranking problem. The risk-based auditing context is in fact an example where even the type of the suitable ranking problem is not determined in advance. One can formulate the problem as a binary ranking problem where the response variable is either tax compliance or a wrong report of the tax liabilities. However, as classification is not as informative as ranking since the classes do not have to be ordered while ranking also incorporates an ordering (see also \cite{furn09}), ranking in turn is less informative than regression since regression tries to predict the actual response values themselves where ranking just tries to find the right ordering. An analogous argument is true for binary ranking problems and continuous ranking problems. If one states a binary ranking problem, one would just get information which taxpayer is most likely to misreport his or her income without providing any information on its amount. On the other hand, if one sets up a continuous ranking problem where the amount of damage is the variable of interest, one can directly get information about the compliance of the taxpayer by looking at the sign of the response value. In particular, if information on the compliance is available, then one can assume that the information on the amount of additional payment or back-payment has also been collected, so imposing a binary ranking problem would lead to a large loss of information. Additionally, the issue that using a regression strategy in order to solve a ranking problem requires stronger assumptions as pointed out in \cite{furn09} does not apply here since the continuous, real-valued responses are the original ones. \ \\

The Boosting-type and most of the SVM-type approaches that we reviewed so far invoke surrogate losses of the hard ranking loss (or even of the 0/1-classification loss). It is discussed in \cite{TWphd} whether an analogous approach is appropriate for a Gradient Boosting algorithm (see e.g. \cite{bu07}) for the hard continuous instance ranking problem. They conclude that due to the support of the response variable which is no longer just $\{-1,1\}$ or some finite set as in the $d-$partite ranking problem, exponential or Hinge surrogates would dramatically fail to be meaningful surrogates for the hard ranking loss. Another weakness would be the necessity to evaluate the gradients of the pair-wise loss (which are sums themselves) in each Boosting iteration, making the algorithm computationally expensive. \ \\

To handle these issues, \cite{TW19c} proposed a so-called ''gradient-free Gradient Boosting'' approach to make Gradient Boosting accessible to non-regular loss functions like the hard ranking loss. Their approach is based on $L_2-$Boosting with component-wise linear baselearners (\cite{bu03}, \cite{bu06}) which minimizes the squared loss by successively selecting the simple linear regression model, i.e., the linear regression model based on one single column, that minimizes the squared loss w.r.t. the resulting combined model most. \cite{TW19c} propose to alternatingly perform $(M-1)$ of these standard iterations for some $M>1$ and one ''singular iteration'' where the linear baselearner which improves the hard ranking loss of the combined strong model most is selected. \ \\

However, they discuss that the resulting Boosting solution suffers from overfitting (as Gradient Boosting solutions without early stopping generally do) and that the predictor set corresponding to the solution is not stable. They argue that a combination with a Stability Selection (\cite{bu10}, \cite{hofner15}) would be necessary which is outlined in \cite{TWphd} where a modified Stability Selection is proposed. This approach, presented in \cite{TWphd}, is the first one that tries to find a stable predictor set for ranking. While the original approach has a complexity of $\mathcal{O}(mn\ln(n)p)$ using Lemma \ref{fasthardrank} to compute the ranking loss in the singular iterations, the aggregation of $B$ such Boosting models in a Stability Selection leads to $B$ times the respective complexity. However, in its current implementation in the $\mathsf{R}-$package \texttt{gfboost}, it is mainly designed for the hard continuous instance ranking problem. Nevertheless, there is no restriction to apply their strategy to bipartite and $d-$partite continuous ranking problems if the underlying $L_2-$Boosting algorithm is replaced by a suitable variant like \texttt{LogitBoost} or \texttt{MultiLogitBoost}. \ \\

The strategy of \cite{furn09}, building up on the work of \cite{furn02}, for multipartite instance ranking problems is not a Boosting-type approach at the first glance, but they essentially combine different weak (classification) models to get a suitable ranking model. Their idea amounts to replacing the multipartite ranking task by a family of binary classification tasks. They consider first an approach going back to \cite{frank} where $(d-1)$ binary problems are defined in the sense that for problem $k$, all instances with a class label in $\{c_1,...,c_k\}$ are interpreted as negative instances all the instances with label in $\{c_{k+1},...,c_d\}$ as positive ones. Then, getting a scoring function $s_k$ for each model, they propose to combine the scores, i.e., to combine the models in contrast to rank aggregation as done in \cite{clem09d}. They decide to sum up the scores $s_k$ to get a final score $s$. Alternatively, \cite{furn09} suggest to learn a binary classification model for each pair $(c_i,c_j)$ of classes and to sum up the individual scores, maybe even in a weighted fashion, to get the final score. As binary classifier, they use logit models. In principle, there is no restriction that prohibits an application of classification algorithms that perform model selection which would lead to the question how to get a suitable aggregated predictor set. 

\subsection{Tree-type approaches} \label{treesec} 

\cite{clem08c}, \cite{clem08d} and \cite{clem09}, for instance, also concentrate on AUC maximization to solve binary instance ranking problems as for example \cite{rak04}, but in a stricter and more sophisticated way. Given the true conditional probability $\eta(x)=P(Y=1|X=x)$ and a scoring function $s$, they introduce metrics on the ROC space which are \begin{center} $ \displaystyle d_1(s,\eta):=\int_0^1 |ROC^*(\alpha)-ROC_s(\alpha)|d\alpha $ \end{center} and \begin{center} $ \displaystyle d_{\infty}(s,\eta):=\sup_{\alpha \in [0,1]}(|ROC^*(\alpha)-ROC_s(\alpha)|) $ \end{center} where $ROC^*$ is the optimal ROC curve and $ROC_s$ the ROC curve induced by the scoring function $s$. Note that the absolute value in the supremum is not necessary since per definition the optimal ROC curve dominates every competitor ROC curve. The idea in the cited references is to optimize the ROC curve according to $d_{\infty}$, i.e., in an $L_{\infty}-$sense due to the disadvantage that an $L_1-$optimiza\-tion is nothing but a AUC-optimization due to \begin{equation*} \begin{gathered} d_1(s,\eta)=\int_0^1 ROC^*(\alpha)d\alpha-\int_0^1 ROC_s(\alpha)d\alpha \\ =AUC^*-AUC_s. \end{gathered} \end{equation*} An AUC-optimization is not appropriate according to the authors since different ROC curves can have the same AUC. \ \\

Cl\'{e}men\c{c}on and coauthors provide tree-type algorithms which turn out to be an impressively flexible class of ranking algorithms that can be applied to all hard instance ranking problems as well as to localized binary instance ranking problems. \ \\

As for binary instance ranking problems, they provided \texttt{Tree\-Rank} and \texttt{RankOver} (\cite{clem08c}, \cite{clem08d}). The idea behind the \texttt{TreeRank} algorithm is to divide the feature space $\mathcal{X}$ into disjoint parts $\mathcal{P}_j$ and to construct a piece-wise constant scoring function \begin{center} $ \displaystyle s_N(x)=\sum_{j=1}^N a_jI(x \in \mathcal{P}_j) $ \end{center} for $a_1>...>a_N$. This results in a ROC curve that is piece-wise linear with $(N-1)$ nodes (not counting $(0,0)$ and $(1,1)$) as shown in \cite[Prop. 13]{clem08c}. The \texttt{TreeRank} algorithm then recursively adds nodes between all existing nodes such that the ROC curve approximates the optimal ROC curve by splitting each region $\mathcal{P}_j$ in two parts. More precisely, one starts with the region $\mathcal{P}_{0,0}=\mathcal{X}$ and the coefficients $\alpha_{0,1}=\beta_{0,1}=1$. In each stage $b=0,...,D-1$ of the tree and in every iteration $k=0,...,2^b-1$, one computes the estimates \begin{center} $ \displaystyle \hat \alpha(\mathcal{P}_{b,k}):=\frac{1}{n_-}\sum_i I(X_i \in \mathcal{P}_{b,k}, Y_i=-1) $ \end{center} \begin{center} $ \displaystyle \hat \beta(\mathcal{P}_{b,k}):=\frac{1}{n_+}\sum_i I(X_i \in \mathcal{P}_{b,k}, Y_i=1) $ \end{center} and optimizes the entropy measure \begin{equation*} \begin{gathered} Ent(\mathcal{P}_{b,k}):= \\ (\alpha_{b,k+1}-\alpha_{b,k})\hat \beta(\mathcal{P}_{b,k})-(\beta_{b,k+1}-\beta_{b,k})\hat \alpha(\mathcal{P}_{b,k}) \end{gathered} \end{equation*} by finding a subset of $\mathcal{P}_{b,k}$ which maximizes this empirical entropy. The coefficients are updated recursively. \ \\

Similarly, the \texttt{RankOver} algorithm constructs a piece-wise linear approximation of the optimal ROC curve by computing a piece-wise constant scoring function, too, but instead of partitioning the feature space, it generates a partition of the ROC space. However, the authors seem to prefer \texttt{TreeRank} over it since their subsequent algorithms are based on the former, so we do not review more details of \texttt{RankOver}.  \ \\

\cite{clem08c} already mention that \texttt{TreeRank} may be used as weak ranker for a Boosting-type approach. \ \\

Extensions by combining the \texttt{TreeRank} algorithm in combination with bagging resp. in a \texttt{RandomForest}-like sense are given in \cite{clem09d}, \cite{clem10b}. A crucial question is how to combine the rankings predicted by the $B$ different trees. This leads to a so-called Kemeny aggregation (\cite{kemeny}, see also \cite{clem17b} for theoretical aspects of rank aggregation) where a consensus ranking is computed. Having some distance measure $D$ which in \cite{clem09d} and \cite{clem10b} may be a Spearman correlation or Kendall's $\tau$, the consensus ranking, represented by a permutation $\pi^* \in \Perm(1:n)$, is the solution of \begin{center} $ \displaystyle \sum_{b=1}^B D(\hat \pi_b, \pi)=\min_{\pi}! $ \end{center} for the predicted permutations $\hat \pi_b$ for tree $b$, respectively. As for the \texttt{RandomForest}-type approach (''Ranking forest''), \cite{clem10b} make two suggestions how to randomize the features in each node. \ \\

As for the pruning of ranking trees, we refer to \cite{clem09e} and \cite{clem10b} who recommend to use the penalized empirical AUC as pruning criterion, i.e., for a tree $T$, one selects the subtree $T_{sub}$ which maximizes \begin{center} $ \displaystyle \widehat{AUC}_{s_{T_{sub}}}-\lambda |T_{sub}| $ \end{center} where $s_T$ denotes the scoring function corresponding to tree $T$. \ \\

The \texttt{TreeRank} algorithm has been available in the $\mathsf{R}-$package \texttt{TreeRank}, but it had been removed. Nevertheless, the source code is still available \footnote{https://github.com/cran/TreeRank}. \ \\

Theoretically, these tree-type algorithms provide an advantage over the algorithms that optimize the AUC since they approximate the optimal ROC curve in an $L_{\infty}-$sense while the competitors just optimize the ROC in an $L_1-$sen\-se (see  \cite[Sec. 2.2]{clem08d}). On the other hand, they suffer from strong assumptions since it is required that the optimal ROC curve is known. Additionally, this optimal ROC curve has to fulfill some regularity conditions which is differentiability and concavity for the \texttt{TreeRank} algorithm and twice differentiability with bounded second derivatives for the \texttt{RankOver} algorithm. \ \\ 

These tree-type algorithms are tailored to bipartite instance ranking problems. However, as pointed out in \cite{clem13}, they can be used for local AUC optimization (see Def. \ref{locaucdef}), so they are applicable for both hard and localized bipartite instance ranking problems while the AUC-maximizing competitors show inferior local ranking performance in the simulation studies of \cite{clem13}. \ \\ \ \\

As for the $d-$partite instance ranking problems, \cite{clem12} build up on the strategy of \cite{furn02} and \cite{furn09} that they can be regarded as collection of bipartite ranking problems if one considers approaches like one-versus-all or one-versus-one. In \cite{clem12}, they apply different algorithms tailored to bipartite ranking problems like \texttt{TreeRank}, \texttt{RankBoost} or \texttt{Ranking\-SVM} and evaluate their performance in the VUS criterion. \ \\

However, since the algorithms are not designed for VUS-optimization, \cite{clem15b} modify their \texttt{TreeRank} algorithm such that the splits of each node are performed first in a one-versus-one sense (but only for adjacent classes) and then the optimal split of them is selected according to the VUS criterion. The resulting \texttt{TreeRankTournament} algorithm therefore is applicable to the hard $d-$partite instance ranking problem. \cite{clem15} provide a bagged and a \texttt{RandomForest}-type version of this algorithm, analogously to the bagged trees for the bipartite case. \ \\ \ \\

Recently, \cite{clem18} provided pioneer work for the hard continuous instance ranking problem which did not have been considered so far. Let w.l.o.g. $Y \in [0,1]$. Then each subproblem \begin{center} $ \displaystyle \max_s(P(s(X)>t|Y>y)-P(s(X)>t|Y<y)) $ \end{center} for $y \in [0,1]$, i.e., $s(X)$ given $Y>y$ should be stochastically larger than $s(X)$ given $Y<y$, is a bipartite instance ranking problem, so the continuous instance ranking problem can be regarded as a so-called ''continuum'' of bipartite instance ranking problems (\cite{clem18}). \ \\

As a suitable performance measure, they provide the area under the integrated ROC curve \vspace{-0.1cm} \begin{center} $ \displaystyle \IAUC(s):=\int_0^1 \IROC_s(\alpha)d\alpha:=\int_0^1 \int \ROC_{s,y}(\alpha)dF_y(y)d\alpha $ \end{center} \vspace{-0.1cm} where $\ROC_{s,y}$ indicates the ROC curve of scoring function $s$ for the bipartite ranking problem corresponding to $y \in ]0,1[$ and where $F_y$ is the marginal distribution of $Y$.  Alternatively, they make use of Kendall's $\tau$ as a performance measure for continuous ranking. \ \\

The approach presented in \cite{clem18} manifests itself in the tree-type \texttt{CRank} algorithm that divides the input space and therefore the training data into disjoint regions. In each step/node, the binary classification problem corresponding to the median of the current part of the training data is formulated and solved. Then, all instances whose predicted label was positive are delegated to the left children node, the others to the right children node. Stopping when a predefined depth of the tree is reached, the instance of the leftmost leaf is ranked highest and so forth, so the rightmost leaf indicates the bottom instance. \ \\

\cite{clem18} already announced a forthcoming paper where a \texttt{RandomForest}-type approach for \texttt{CRank} will be presented. \ \\

All these tree-type approaches focus on a sophisticated optimization of the AUC or another appropriate criterion. For the price of getting models that are difficult to interpret, these techniques are very flexible and are applicable to the most types of instance ranking problems.

\subsection{Approaches with neural networks and Deep Learning} 

\cite{burges} suggest to define a pair-wise variant of the cross-entropy loss as surrogate for the hard ranking loss. More precisely, their pair-wise cross-entropy loss is given by \begin{center} $ \displaystyle L^{CE}_{ij}(s):=-p_{ij}\ln(\hat p_{ij})-(1-p_{ij})\ln(1-\hat p_{ij}) $ \end{center} where \begin{center} $ \displaystyle \hat p_{ij}:=\frac{\exp(\hat s(X_i)-\hat s(X_j))}{1+\exp(\hat s(X_i)-\hat s(X_j))} $ \end{center} and where $p_{ij}$ is the analog for the theoretical differences. From a probabilistic point of view, the $p_{ij}$ are interpreted as posterior probabilities that instance $i$ is ranked higher than instance $j$. The main contribution of \cite{burges} is to generalize the back-propagation algorithm used when fitting neural networks. \ \\

They propose a two-layer neural network and define the following pair-wise linear combination of features: \begin{equation*} \begin{gathered} s(X_i):= \\ h^{(3)}\left(\sum_j w_{ij}^{(32)}h^{(2)}\left(\sum_k w_{jk}^{(21)}X_k+b_j^{(2)}\right)+b_i^{(3)}\right). \end{gathered} \end{equation*} The $h^{(l)}$ are considered to be activation functions. The back-propagation algorithm then is based on the partial derivatives of $s$ w.r.t. the weights resp. the offsets. \ \\

Again, this \texttt{RankNet} algorithm is tailored to the hard bipartite instance ranking problem and the experiments in \cite{burges} are based on document retrieval data. It is available at the RankLib library (\cite{ranklib}). \ \\

\cite{song16} introduce an approach based on gradients of the expected loss. Their work is based on \cite{hazan10} who proved that \begin{equation*} \begin{gathered} \nabla_{\theta}\E[L(Y,s_{\theta}(X))]= \\ \lim_{\epsilon \rightarrow 0}\left(\frac{1}{\epsilon}\E[\nabla_{\theta}F(X,Y_{direct},\theta)-\nabla_{\theta}F(X,Y_{\theta},\theta)]\right)  \end{gathered} \end{equation*} where \begin{center} $ \displaystyle Y_{direct}=\argmax_{\tilde Y}(F(X,\tilde Y,\theta) \pm \epsilon L(Y,\tilde Y)) $ \end{center} and \begin{center} $ \displaystyle Y_{\theta}:=\argmax_{\tilde Y}(F(X,\tilde Y,\theta)) $ \end{center} for some function $F$ that is linear in $\theta$. \cite{song16} extend these results for non-linear and non-convex functions. \ \\

In fact, \cite{song16} apply their results to bipartite hard instance ranking problems by setting \begin{equation*} \begin{gathered} F(X,Y,\theta):=\\ \frac{1}{n_-n_+}\sum_{i: Y_i=1} \sum_{j: Y_j=-1} r(X_i,X_j)(\hat s_{\theta}(X_i)-\hat s_{\theta}(X_j)) \end{gathered} \end{equation*} for the ranking rule introduced in Def. \ref{rankruledef} and by invoking the loss function \begin{equation*} \begin{gathered} L(Y,\hat Y):=\\ 1-\frac{1}{n_+}\sum_{j: \rk(\hat Y_j)=1} \frac{1}{n_+} \sum_i I(\rk(\hat Y_i) \le j)I(Y_i=1) \end{gathered} \end{equation*} where $\hat s_{\theta}$ is the scoring function that is learned by the Deep Neural Network. \cite{song16} prove how their theoretical results can be applied to the case with the given functions $F$ and $L$ and show that a back-propagation strategy with a suitable Bellman recursion is available. \ \\

\cite{engilberge} propose to use Deep Learning and essentially combine two Deep Neural Networks. They discuss several smooth surrogate losses, for example for losses corresponding to Spearman correlation, Mean Average Precision or Recall and argue that since they are all rank-based, i.e., depend on $\rk(Y)$ and $\rk(\hat Y)$, it is hard to optimize them due to non-differentiability. Therefore, they propose to invoke a real-valued scoring function such that  the fitted scoring function $\hat s_1$ approximates the true ranking vector $\rk(Y)$ as best as possible by considering the $L_1-$loss function \begin{center} $ \displaystyle \frac{1}{n} ||s_1(X_i)-\rk(Y_i)||_1. $ \end{center} According to \cite[Sec. 3.2]{engilberge}, $\hat s_1$ needs to be trained on synthetic training data, using a sorting Deep Neural Network.  \ \\

Having real-valued scores, they propose the surrogate loss \begin{center} $ \displaystyle \sum_i ||\hat s_1(s_2(X_i))-\rk(Y_i)||_2^2 $ \end{center} for a loss based on Spearman's correlation and in the case of multilabel responses with classes $1,...,d$, they propose the surrogate loss \begin{center} $ \displaystyle \sum_{k=1}^d \langle \hat s_1(s_2(Y)_k), Y_k \rangle $ \end{center} based on the Mean Average Precision, where $Y_k$ is a binary vector with ones where the respective component of $Y$ is from class $k$ and where $s_2(Y)_k$ is considered to be the score vector for class $k$. They also propose a surrogate for the Recall criterion. The scoring function $\hat s_2$ is again computed using a Deep Neural Network. \ \\

\cite{engilberge} call their approach \texttt{SoDeep} and apply it to media memorability, image classification and cross-modal retrieval tasks, each task corresponding to one of their three surrogate losses. In fact, \texttt{SoDeep} is applicable to hard and localized (the latter with the surrogate for Recall) ranking problems and does not make requirements for $\mathcal{Y}$. On the other hand, the solution, as well as the one from \cite{song16}, suffers from the common disadvantages of Deep Neural Networks, i.e., they do not perform variable selection and are very difficult to interpret. 

\section{Discussion} \label{discuss}

\subsection{Discussion of the different ranking problems}\label{rankvsord} 

In this subsection, we discuss the different types of ranking problems introduced earlier from a qualitative point of view and the differences between ranking and ordinal regression. \ \\

Ordinal regression problems are indeed very closely related to ranking problems. As already pointed out in \cite{robbiano}, especially multipartite ranking problems (\cite{clem12}) share the main ingredient, i.e., the computation of a scoring function that should provide pseudo-responses with a suitable ordering. However, the main difference is that the multipartite ranking problem is already solved once the ordering of the pseudo-responses is correct while the ordinal regression problem still needs thresholds such that a discretization of the pseudo-responses into the $d$ classes of the original responses is correct, see also \cite{furn09}. \ \\

Note that, on the other hand, due to the discretization, ordinal regression problems can indeed be perfectly solved even if the rankings provided by the scoring function are not perfect. For example, consider observations with indices $i_1,...,i_{n_k}$ that belong to class $k$. If for a scoring rule $s$ we had the predicted ordering $s(X_{i_1})<s(X_{i_2})<...<s(X_{i_{n_k}})$ but the true ordering is different, then we can still choose thresholds such that all $n_k$ instances that belong to class $k$ (and no other instance) are classified into this class, provided that $s(X_i) \notin [s(X_{i_1}),s(X_{i_{n_k}})] \ \forall i \notin \{i_1,...,i_{n_k}\}$. \cite{furn09} argued that the class labels could in principle also be used as ranking scores but that this strategy would clearly lead to many ties. Though, as \cite{robbiano} already pointed out, the ordinal regression is based on another loss function. \ \\

Concerning informativity, one can state that multipartite ranking problems are more informative than ordinal regression problems due to the chunking operation that is performed for the latter ones. But in fact, in an intermediate step, i.e., when having computed the scoring function, the ordinal regression problem is as informative as multipartite ranking problems. This is also true for standard logit or probit models (the two classes generally are not ordered, but when artificially replacing the true labels by $-1$ and $+1$ where the particular assignment does not affect the quality of the models, they can at least mathematically be treated as ordinal regression models) where the real-valued pseudo-responses computed by the scoring function are discretized at the end to have again two classes. As for informativity, see also \cite{furn09} who point out that classification is less informative than ordinal regression (and therefore ranking) but that regression may  make too strong assumptions like requiring that one can compute meaningful differences between the numerical class values.  \ \\

A similar discussion can be found in \cite{huller10b} and \cite{huller10c} in the context of label ranking. In contrast to classification where a model picks one of the labels, the goal in label ranking is to predict an ordering on the label set for each instance. They point out that a classification model predicts the most probable class but sorting the labels according to the predicted probabilities that a particular instance belongs to the respective class similarly induces a ranking on the label set. \ \\

The continuous instance ranking problem can be treated as a special case where no pseudo-responses are needed since the original responses are already real-valued, but again, instead of optimizing some regression loss function, the goal is actually to optimize a ranking loss function. \ \\ 

For further discussions on the relation of ranking and ordinal regression (also called ''ordinal classification'' and ''ordinal ranking'' in the reference), see \cite{lin08}. \ \\ 

From this point of view, the three combined problems for the continuous case, i.e., weak, hard and localized continuous instance ranking problems\index{Ranking problems!Continuous|)}, are easy to distinguish and are all meaningful. Hard bipartite and hard $d-$partite instance ranking problems\index{Ranking problems!Hard|)} are essentially solved by most of the algorithms that we described in Section \ref{algsec} and localized bipartite ranking problems can be solved using the tree-type algorithms of Cl\'{e}men\c{c}on as pointed out for instance in \cite{clem13}. Clearly, these localized bipartite problems directly reflect the motivation from risk-based auditing or document retrieval. \ \\

It has been mentioned in \cite{clem15b} that their tree-type algorithm is not able to optimize the VUS locally. To the best of our knowledge, this has not been achieved until now. But indeed, localized $d-$partite ranking problems\index{Ranking problems!Localized|)} can also be interesting in document retrieval settings where the classes represent different degrees of relevance. Then it would be interesting for example to just recover the correct ranking of the relevant instances, i.e., the ones from the ''best'' $(d-1)$ classes if class $d$ represents the ''rubbish class''.  \ \\

As mentioned earlier, weak ranking problems can be identified with binary classification with a mass constraint (\cite{clem08b}). In the case of weak bipartite instance ranking problems\index{Ranking problems!Bipartite|)}, it may sound strange to essentially mix up two classification paradigms, but one can think of performing binary classification by computing a scoring function and by predicting each instance as element of class 1 whose score exceeds some threshold, as it is done for example in logit or probit models. One can think of choosing the threshold such that there are exactly $K$ instances classified into class 1 instead of optimizing the AUC or some misclassification rate. \ \\

The only combination that does not seem to be meaningful at all would be weak $d-$partite ranking problems. By its inherent nature, a weak ranking problem imposes are binarity which cannot be reasonably translated to the $d-$partite case. Even in the document retrieval setting, a weak $d-$partite ranking\index{Ranking problems!$d-$partite|)}\index{Ranking problems!Weak|)} problem may be thought of trying to find the $K$ most important documents which implied that the information that is already given by the $d$ classes would be boiled down to essentially two classes, so this combination is not reasonable.

\subsection{Relation to other ranking problems} 

We already distinguished between instance, object and label ranking problems and restricted ourselves to instance ranking problems. However, there is one interesting approach where the algorithms that we reviewed in this work for instance ranking are applicable for object ranking, i.e., unsupervised ranking. \ \\
 
Usually, one invokes a very popular probabilistic approach for object ranking, i.e., to predict a probability distribution on the set $\Perm(1:n)$. Two prominent models are the Mallows model and the Plackett-Luce model. The Mallows model (\cite{mallows}) is based on distances between different permutations, in general based on Kendall's $\tau$, which leads to a maximum likelihood approach. The Plackett-Luce model (\cite{luce}, \cite{plackett}) performs a Bayes estimation. These models have successfully entered object ranking (\cite{huller15}) and label ranking (\cite{huller14}, \cite{cheng09}). \ \\   

Being inherently different from instance ranking problems, there indeed exists a paradigm which relates object ranking problems to instance ranking problems. \cite{huller17} point out that when using a scoring function approach in object ranking, i.e., ranking $x$ before $x'$ if $s(x)>s(x')$, and if the training data are of the form $(X^{(k)},\pi^{(k)})_{k=1}^K$ for sets $X^{(k)}$ of objects and corresponding true orderings $\pi^{(k)} \in \Perm(1:|X^{(k)}|)$, one may follow a pointwise resp. a pairwise paradigm. While the pointwise paradigm means to replace each object with a set of labeled instances where the labels depend on the permutation value and the size of the actual set $X^{(k)}$ (see \cite{kamishima} for this so-called expected rank regression approach), the pairwise approach aims to learn all pairwise preferences where $X_{\pi^{-1}(i)}$ is preferred over $X_{\pi^{-1}(j)}$ for $i<j$ which translates the object ranking problems to a set of binary classification problems where the preferred object is interpreted as the object of the positive class. This reasoning lets instance ranking algorithms for bipartite ranking enter object ranking, of which \cite{huller17} use \texttt{RankingSVM}. \ \\

Evidently, one may also distinguish between hard, weak and localized object and label ranking problems. Such an idea has already been proposed in \cite{furn08} for label ranking where one has to learn both a (hard) ranking but also a binary classification into relevant and non-relevant labels.

\section{Conclusion and outlook}

We provided a systematic review of different ranking problems, concerning both the type of the response variable and the goal of the analyst. We analyzed and discussed the corresponding loss functions resp. quality criteria and carefully discussed different types of instance ranking problems and distinguished instance ranking problems from object and label ranking problems.  \ \\

Section \ref{algsec} contains a detailed review of existing learning algorithms for instance ranking based on the empirical resp. structural risk minimization principle in a unified notation, grouped by the underlying machine learning algorithm.

\subsection{Open problems} 

Despite there exists a vast variety of approaches to solve instance ranking problems, most of the current approaches are either designed for discrete- or for continuous-valued response variables. Additionally, nearly all of the reviewed techniques require an appropriate surrogate loss function for one of the ranking losses which is generally convex and therefore cannot be regarded as robust in the sense of robust statistics (e.g. \cite{huber}, \cite{hampel}). Tree-type and Deep Learning approaches usually suffer from the lack of interpretability. Similarly, many of the approaches do not invoke a suitable sparse model selection. \ \\

As for future research, a unified approach which does not depend on whether the response variable is categorical or continuous and which provides a sparse, robust, stable and well-interpretable model would be a desirable goal. Deep Learning has gained a lot of attention during the last decade as is capable to result in excellent predictions, but interpretability of the model is still an ongoing research question. \ \\

An even more difficult situation arises once the response variable is multivariate, i.e., one has $\mathcal{Y} \subset \R^k$, $k \ge 2$. Then one can clearly get partial rankings (not to be confused with partial orders in \cite{cheng10b} which reflect the uncertainty that prohibit a clear ordering) which are rankings for each column of $Y$ separately. However, since one is actually interested in the ranking of the rows $X_i$ which in the case of univariate responses just equals the ranking of the $Y_i$, it remains to find an overall ranking for the $X_i$ in the case that each response column corresponds potentially to a different ranking. There are many situations where one has partial rankings and wants to get a suitable combined ranking based on these partial rankings. Such situations range from the ranking of websites by different search engines (\cite{dwork}) to the combination of judge grades in competitions (\cite{davenport}) and even to applications in nanotoxicology (\cite{patel}). The aggregation of the partial rankings gets even more difficult if the quality of the partial rankers is different (\cite{deng14}). A standard approach is to compute an consensus ranking using for example the Kemeny aggregation (\cite{kemeny}). However, if additionally sparse (and stable) model selection is desired, one has to find a suitable predictor set w.r.t. all response columns which for regression has already been done by \cite{lutz}). A first idea to solve this problem has been outlined in \cite{TWphd}.



\renewcommand\refname{References}
\bibliography{Biblio}
\bibliographystyle{abbrvnat}

\end{small}

\end{document}